\documentclass[runningheads]{llncs}

\usepackage{eccv}

\usepackage{bm}
\usepackage{adjustbox}

\usepackage{eccvabbrv}

\usepackage{graphicx}
\usepackage{booktabs}

\usepackage[accsupp]{axessibility}  

\usepackage{wrapfig, blindtext} 
\usepackage{multirow}

\usepackage[pagebackref,breaklinks,colorlinks,citecolor=eccvblue]{hyperref}

\usepackage{orcidlink}

\begin{document}
\title{Seeing the Unseen: A Frequency Prompt Guided Transformer for Image Restoration} 


\titlerunning{A Frequency Prompt Guided Transformer for Image Restoration}

\author{Shihao Zhou\inst{1,2} \and
Jinshan Pan\inst{3} \and
Jinglei Shi\inst{1} \and 
Duosheng Chen\inst{1}\and \\
Lishen Qu\inst{1}\and
Jufeng Yang\inst{1,2} }

\authorrunning{Zhou et al.}

\institute{VCIP \& TMCC \& DISSec, College of Computer Science, Nankai University \and
Nankai International Advanced Research Institute (SHENZHEN· FUTIAN) \and
School of Computer Science and Engineering, Nanjing University of Science and Technology}

\maketitle

\begin{abstract}
%

How to explore useful features from images as prompts to guide the deep image restoration models is an effective way to solve image restoration. 
In contrast to mining spatial relations within images as prompt, which leads to characteristics of different frequencies being neglected and further remaining subtle or undetectable artifacts in the restored image, we develop a \textbf{F}requency \textbf{Pro}mpting image restoration method, dubbed \textbf{FPro}, which can effectively provide prompt components from a frequency perspective to guild the restoration model address these differences. 
%
%
%
%
%
Specifically, we first decompose input features into separate frequency parts via dynamically learned filters, where we introduce a gating mechanism for suppressing the less informative elements within the kernels. 
To propagate useful frequency information as prompt, we then propose a dual prompt block, consisting of a low-frequency prompt modulator (LPM) and a high-frequency prompt modulator (HPM), to handle signals from different bands respectively. 
Each modulator contains a generation process to incorporate prompting components into the extracted frequency maps, and a modulation part that modifies the prompt feature with the guidance of the decoder features. 
Experimental results on commonly used benchmarks have demonstrated the favorable performance of our pipeline against SOTA methods on 5 image restoration tasks, including deraining, deraindrop, demoir{\'e}ing, deblurring, and dehazing. 
{The source code and pre-trained models will be available at \href{https://github.com/joshyZhou/FPro}{https://github.com/joshyZhou/FPro}}.
  
\end{abstract}

\section{Introduction}
\label{sec:intro}
%
%
%
Capturing images in unsatisfactory environments, e.g., rain, haze, usually leads to low-quality ones that accordingly affect the application of downstream tasks in practice. Thus, developing an effective image restoration method to restore clear images from degraded ones is an important task.

%

Significant progress has been made due to kinds of the deep learning models~\cite{chen2022simple,zamir2022restormer,chen2021pre}, and these deep learning-based approaches become predominant ones as they achieve better performance than the conventional hand-crafted prior-based  approaches~\cite{zhang2021pami,he2010single,chantas2009variational,narasimhan2003contrast,li2016rain}.

%
%

{}{
Existing methods, e.g.,~\cite{wang2022uformer,iccv2021_swinIR,zamir2022restormer} achieve promising performances in kinds of image restoration tasks. 
However, these learning-based methods intend to learn a mapping function between degraded images and clear ones, where the characteristics of the specific degradation are less considered. 
For example, rain streaks tend to obscure the background partially, whereas raindrops typically result in a more pronounced regional occlusion. 
Accordingly, these models are hindered from generating better results. 
}
%

%
%
More recently, prompt-learning based methods~\cite{wang2023promptrestorer,potlapalli2023promptir,wang2023selfpromer} serve as an alternative approach to encode useful content of specific degradation for modulating the network
{}{, and make a clear performance boost for image restoration.}
%
However, we notice that these methods~\cite{potlapalli2023promptir,wang2023promptrestorer} pay attention to mining spatial correlations to provide degradation information, whereas the task-specific frequency cues are less studied. 
%
%
%
%
%
%
{}{
Indeed, since various forms of degradation exhibit distinct impacts on image content, they affect information from different frequency bands. 
}
Hence, it is crucial to develop an efficient prompt mechanism that explores useful prompts from a frequency perspective for identifying specific characteristics of diverse degradation, which can boost the model to effectively restore images with finer details and non-local structures of the scenes. 

%
%
This paper proposes a \textbf{F}requency \textbf{Pro}mpting image restoration method, dubbed \textbf{FPro}, to {}{modulate the network by encoding degradation-specific frequency cues as prompts. }
%
%
%
%
{}{As mentioned above, existing prompt strategies~\cite{wang2023promptrestorer,potlapalli2023promptir} focus on mining spatial relations as useful prompts. }
%
{}{In this way, differences between the restored image and the real one within frequency domain~\cite{jiang2021focal} are ignored, which remain subtle or undetectable artifacts in the spatial domain.} 
Instead, our FPro aims to enjoy benefits from the capability of prompt learning in different frequency bands at multi-scale resolutions to recover clean images. 

We present two designs to make FPro suitable for image restoration: 
%
1). We first decouple input features into separate low-/high-frequency parts using a gated dynamic decoupler, {}{as signals in different frequency bands encode image patterns from distinct views, \ie, local details and global structures. }
%
To this end, a gating mechanism is introduced to help learn the enhanced low-pass filters by suppressing the less informative elements within the kernel, which are then employed to generate low-frequency maps. 
Meanwhile, the corresponding high-pass filter is obtained by subtracting the low-pass filter from the identity kernel, for generating high-frequency maps. 
2). We propose a Dual Prompt Block~(DPB), which consists of two modulators, \ie, the Low-frequency Prompt Modulator~(LPM) and the High-frequency Prompt Modulator~(HPM), to handle low- and high-frequency {information} respectively. 
Each {modulator} includes (a) {a generation part that} incorporates prompting components into the extracted frequency maps{}{, which is supposed to help distinguish various elements within features, such as rain patterns in the context of deraining}; and (b) a modulation part that modifies the prompt feature with the guidance of the feature in the restoration process. 
{In terms of functionality}, LPM enhances the low-frequency characteristics through a gating mechanism in the Fourier domain before injecting the prompting components, which is proven {equivalent to} dynamic large-kernel depth-wise convolution in the spatial domain while computationally efficient, and then encodes low-frequency interactions via global cross-attention. 
{As a complement}, HPM applies a locally-enhanced gating mechanism to obtain useful high-frequency signals, and then encodes high-frequency interactions via local cross-attention. 

Our main contributions in this paper can be summarized as follows:

\begin{itemize}
\item 
We propose FPro, which benefits from prompting learning of frequency components for general image restoration. 
{}{Instead of mining spatial relations as in previous methods, we explore frequency maps to encode specific degradation information as prompts to guide the image restoration model for restoring finer details and the global structure of the scenes.}
\item 
We decouple input features into different frequency bands using learnable low-pass filters, and propose a dual prompt block, which {is composed of} low-frequency prompt modulator (LPM) and high-frequency prompt modulator (HPM), to explore both details and structures for better restoration.
%
\item 

{Experimental results on several image restoration tasks, including deraining, deraindrop, demoir{\'e}ing, deblurring and dehazing, show that FPro achieves favorable performance, compared to state-of-the-art methods.}
\end{itemize}


\section{Related Work}
\label{sec:relatedWork}
\noindent\textbf{Image Restoration.}
Image restoration aims to recover high-quality images from the degraded version. 
Going beyond conventional prior-based solutions~\cite{he2010single,chantas2009variational}, this community has witnessed the great success of a body of learning-based approaches~\cite{yuke_pami22_car_blur_noise,liu2018non,pami22_panXinggang}. 
Despite the promising results obtained by various CNN-based architectures~\cite{pami22_blur_renWenqi,pami21_blur_liujun,cvpr2021cho}, the main concern for methods of this kind is that they pose a limited receptive field problem of the basic convolution operation. 
This means that the feature map contains less global context (corresponding to low-frequency characteristics in an image), and the final prediction can get stuck in this limitation. 
This drawback has motivated the increased interest in exploring components to capture desired global cues, like attention mechanisms~\cite{pami21_deng,tip23_songxibin,niu2020single}, where better restoration performance can be achieved. 
For instance, MIRNet~\cite{zamir2020learning} proposes a dual attention unit to capture contextual information in dual dimensions. 
NLSN~\cite{sr_localSparseAtt_cvpr21} employs a self-attention mechanism to collect global correlation information for super-resolution. 

\noindent\textbf{Transformer-based Restoration.}
The idea of using Transformer architecture~\cite{vaswani2017attention} to address various computer vision tasks has been popular in recent years. 
Thanks to their discriminative feature representation capability, they not only earn advantages in solving high-level vision tasks~\cite{iclr2021_vit,wang2021pyramid,d2021convit}, but also are extended to low-level image restoration tasks~\cite{zhao2023comprehensive,DRSformer,kong2023efficient}. 
Unfortunately, as vanilla self-attention has quadratic complexity to the image size, this mechanism suffers from non-trivial computational costs in handling high-resolution input. 
To address this, some attempts have been made to explore efficient transformer architectures~\cite{chen2021pre,wang2022uformer,zheng2022cross}. 
Specifically, SwinIR~\cite{iccv2021_swinIR} introduces a window-based self-attention scheme to improve efficiency. 
Restormer~\cite{zamir2022restormer} adopts channel-wise self-attention to reduce the computational costs. 
The majority of these works have offered reliable solutions to recover clean images, however, some works~\cite{park2022how,Dong_2023_ICCV} realized that the low-pass filter nature of self-attention, which could lose the high-frequency information, such as textures and edges. 
Even though these models have achieved superior performance, few high-frequency details can be leveraged to implement image restoration, limiting better recovery as a result. 

\noindent\textbf{Visual Prompt Learning.}
More recently, the emergence of prompt learning~\cite{nips20_LLMisfewshotlearner} in natural language processing has resulted in rapid progress in adapting it to vision-related tasks~\cite{Khattak_2023_CVPR,eccv22_VisualPromptTuning,aaai23_visualDomainPrompt}. 
Contrary to high-level vision problems, motivated by high effectiveness, some works also consider seeking the right prompt for the low-level vision models~\cite{ma2023prores,yu2024scaling,wu2023seesr}. 
%
%
%

The goal of this work is not to explicitly prompt the model with the specific degradation type for addressing the ALL-in-One problem (in fact, the previous works of \cite{ma2023prores,li2023prompt,potlapalli2023promptir} have addressed this nicely by designing various degradation prompt modules). 
%
However, our approach is relevant to recent studies~\cite{wang2023promptrestorer,wang2023selfpromer} exploring degradation-specific information for better image restoration results.
In contrast to these attempts that generate raw degradation features with a pre-trained model, we propose to prompt the restoration models from a frequency perspective. 
%
By discerning high-frequency details information and low-frequency global characteristics as 
prompts, our model benefits from information within these frequency bands crucial for addressing degradations. 
This tailored extraction ensures that the model hones in on specific image characteristics directly related to the restoration task. 
%


\section{Proposed Method}
%
\begin{figure}[t]
\centering
\includegraphics[width=\linewidth]{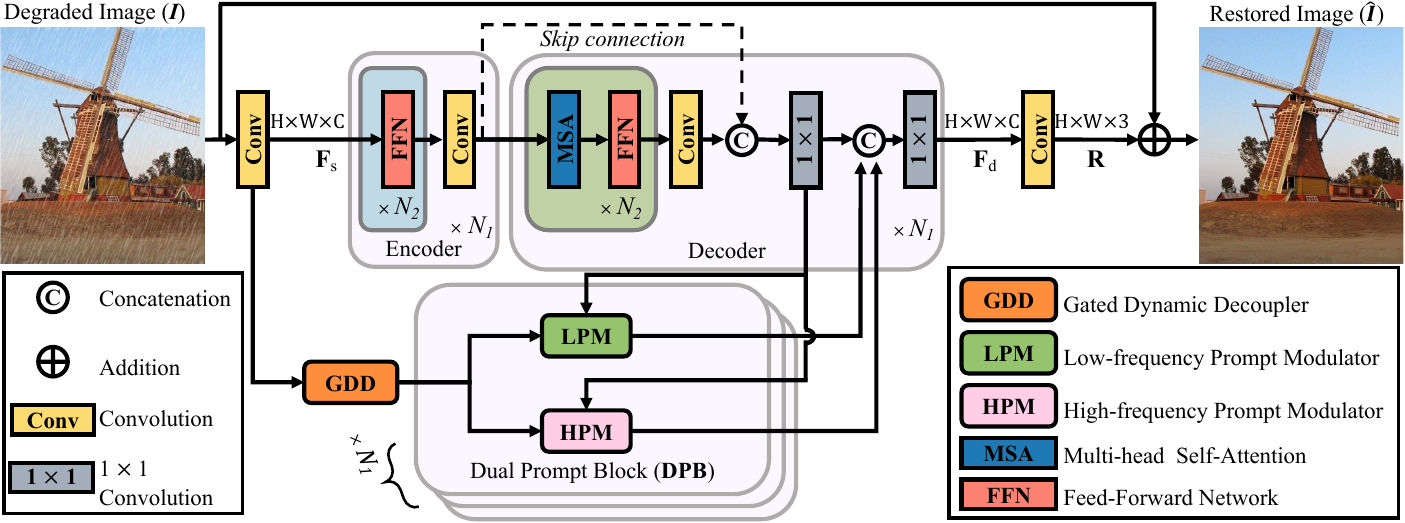}
\vspace{-2.5mm}
\caption{Overview of the proposed \textbf{FPro}. 
Except for the common upper restoration branch, which is similar to existing methods~\cite{iccv2021_swinIR,zamir2022restormer}, FPro contains another bottom prompt branch to extract {}{informative features} 
from a frequency perspective. 
Specifically, the primary components of the prompt branch in this framework are the gated dynamic decoupler (GDD) and dual prompt block~(DPB). 
The GDD is employed to decompose the low-frequency components and corresponding high-frequency characteristics from the input features. 
Then these frequency-specific features are further processed in DPB, \ie, the high-frequency prompt modulator~(HPM) and the low-frequency prompt modulator~(LPM), which generates representative frequency prompt to facilitate the clear image reconstruction. 
}
\label{pic:overall}
\vspace{-2.5mm}
\end{figure}
\subsection{Overall Pipeline}
As depicted in Fig.~\ref{pic:overall}, the overview of our proposed FPro contains the upper restoration branch, like existing works~\cite{iccv2021_swinIR,zamir2022restormer}, and the bottom prompt branch to {}{extract informative frequency maps and then modulate them as prompts.} 
%
\noindent\textbf{Restoration Branch.}
Given a degraded image $\mathbf{I}\in\mathbb{R}^{{H}\times{W}\times3}$ {}{as input}, FPro first applies a convolution layer to extract shallow feature $\mathbf{F}_s\in\mathbb{R}^{{H}\times{W}\times{C}}$; where $H\times{W}$ represents the spatial dimension and $C$ is the {}{number of channel}. 
Next, the shallow feature passes through the upper $N_1$-level encoder-decoder restoration branch to extract deep feature $\mathbf{F}_d\in\mathbb{R}^{{H}\times{W}\times{C}}$. 
%
%
{}{Early layers in Transformer-based models focus on aggregate local patterns~\cite{NEURIPS2021_Early_Convs}, whereas the self-attention module acts as a low-pass filter and tends to dilute high-frequency local details~\cite{park2022how}. 
To alleviate the two contradictory factors, we remove the attention mechanism within the encoder of the restoration branch. 
}
%
Specifically, each level of the encoder includes $N_2$ feed-forward network (FFN)~\cite{zamir2022restormer}  
and the paired convolution layer for down-sampling. 
%
%
The encoder features are fused with the decoder features via skip connections by $1\times{1}$ convolution. 
For the decoder part, each level is composed of $N_2$ pairs of FFN and multi-head self-attention mechanisms (MSA)~\cite{zamir2022restormer}, along with the convolution layer for up-sampling. 
%
%
Finally, a $3\times{3}$ convolution layer is employed to deep feature $\mathbf{F}_d$ for generating residual image $\mathbf{R}\in\mathbb{R}^{{H}\times{W}\times{3}}$. 
The restored image $\hat{\mathbf{I}}$ is estimated by: $\hat{\mathbf{I}}=\mathbf{I}+\mathbf{R}$. 

\noindent\textbf{Prompt Branch.}
{}{In this branch, we take as input the shallow feature $\mathbf{F}_s$ to generate useful frequency prompts, which are further leveraged to facilitate the latent clear image reconstruction. 
%
To achieve this goal, we first decompose the input feature into different frequency bands using a gated dynamic decouple (GDD) (see Section~\ref{subsec:EDD}). 
After that, low-/high-frequency maps are injected with prompt components to distinguish informative elements according to specific tasks, and then modulated as different prompts (\ie, $\mathbf{F}^{out}_{hi}$ and $\mathbf{F}^{out}_{low}$) to interact with the decoder features by $1\times{1}$ convolution (see Section~\ref{subsec:DPB}). 
%
%
%
Next, we present the modules of the prompt branch.
}


\subsection{Gated Dynamic Decoupler}
\label{subsec:EDD}
\begin{figure}[t]
\centering
\includegraphics[width=\linewidth]{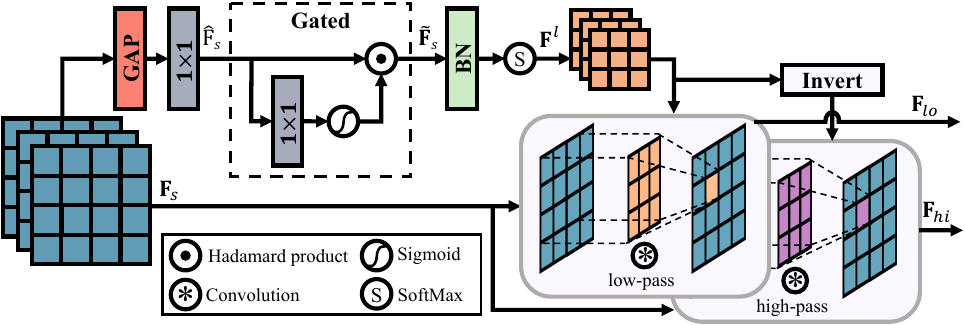}
\vspace{-2.5mm}
\caption{Illustrations of the Gated Dynamic Decoupler. 
}
\label{pic:component_EDD}
\vspace{-2.5mm}
\end{figure}

Each type of degradation affects image content in different ways. 
For instance, rain streaks partially occlude the background while raindrops often cause much greater obstruction, which corresponds to touch high-/low-frequency bands respectively. 
To handle these differences, as shown in Fig.~\ref{pic:component_EDD}, we decompose the input features into separate frequency parts based on gated and dynamically learned filters. 
%
The key ingredient is to introduce a gating mechanism to help generate the gated learnable low-pass filter and the corresponding high-pass filter, which are then employed to obtain low- and high-frequency maps, respectively. 
These filters are dynamically learned for each spatial location and channel group to balance computation burden and feature diversity. 
Specifically, given the input shallow feature map $\mathbf{F}_s\in\mathbb{R}^{{H}\times{W}\times{C}}$, we firstly predicts the low-pass filter for each feature channel group, which can be formulated as:
\begin{equation}
\begin{aligned}
        \hat{\mathbf{F}}^{}_{s} &={\rm Conv}_{1\times 1}({\rm GAP(\mathbf{{F}}_{s}))} ,\\
        \tilde{\mathbf{F}}^{}_{s} &=\hat{\mathbf{F}}^{}_{s}\odot\phi({\rm Conv}_{1\times 1}(\hat{\mathbf{F}}^{}_{s})) ,\\
        {\mathbf{F}}^{l}&={\rm Softmax}(\mathcal{B}(\tilde{\mathbf{F}}^{}_{s} ))
\label{eq:low-pass_gene}
\end{aligned}
\end{equation}
where ${\mathbf{F}}^{l}\in\mathbb{R}^{g\times{k}^2\times{1}\times{1}}$, $g$ is the number of channel groups and $k^2$ corresponds to the kernel size of the learned filter; ${\rm GAP(\cdot)}$ and ${\rm Conv}_{1\times 1}(\cdot)$ are global average pooling layer and convolution operation with the filter size of $1\times1$, respectively; $\phi(\cdot)$ denotes sigmoid activation, $\odot$ refers to the Hadamard product, and $\mathcal{B}(\cdot)$ means Batch Normalization. 
Particularly, ${\rm Softmax}(\cdot)$ is a softmax layer, which ensures the generated filters are low-pass~\cite{zou2023delving}. 
Then, we apply these learned filters to each group input feature $\mathbf{{F}}^{}_{i}\in\mathbb{R}^{{H}\times{W}\times{C_i}}$ to obtain low-frequency components:
\begin{equation}
\begin{aligned}
        \mathbf{F}^{lo}_{i,c,h,w} = \sum\limits_{p,q}^{} {\mathbf{F}}^{L}_{i,p,q}\mathbf{{F}}^{}_{i,c,h+p,w+q},
\label{eq:low-pass_apply}
\end{aligned}
\end{equation}
where ${\mathbf{F}}^{L}\in\mathbb{R}^{{g}\times{k}\times{k}}$ is the reshaped filter, $i$ denotes the group index, $C_i = \frac{C}{g}$ refers to number of the group channel, $c$ means the index of a channel, $h$ and $w$ are spatial coordinates, $p,q\in\{-1,0,1\}$ point to the surrounding locations. 

Meanwhile, we invert this process by subtracting the low-pass filter from the identity kernel to attain the high-pass filter, which is employed to generate the corresponding high-frequency components $\mathbf{F}_{hi}$. 

\subsection{Dual Prompt Block}
\label{subsec:DPB}
Considering that the extracted features, \ie, low-/high-frequency maps, encode image patterns from distinct views (local detail and main structure of the image). 
We design the Dual Prompt Block that includes two components, \ie, High-frequency Prompt Modulator (HPM) and Low-frequency Prompt Modulator ({LPM}), to deal with these feature maps, respectively. 
%

\begin{figure}[t]
\centering
\includegraphics[width=\linewidth]{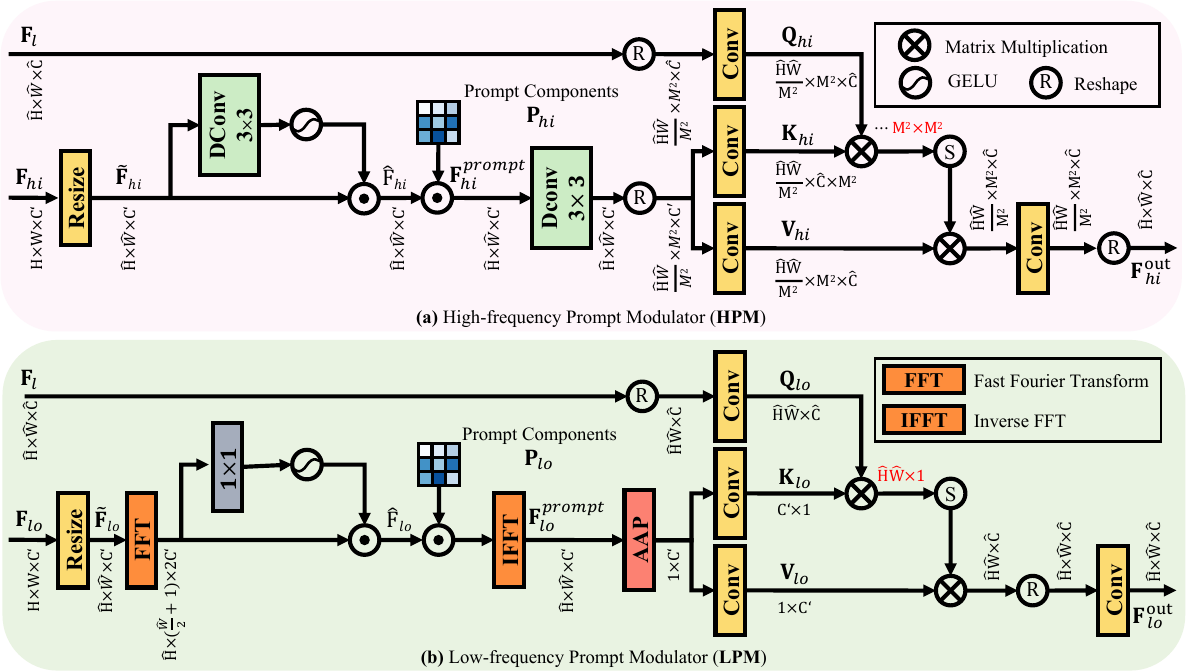}
\vspace{-2.5mm}
\caption{Illustrations of the proposed components. 
(a) High-frequency Prompt Modulator (\textbf{HPM}); (b) Low-frequency Prompt Modulator (\textbf{LPM}). 
}
\label{pic:components}
\vspace{-1.5mm}
\end{figure}

\noindent\textbf{High-frequency Prompt Modulator.}
Given the two input feature maps, including the $l$-level feature $\mathbf{F}_l\in\mathbb{R}^{\hat{H}\times\hat{W}\times\hat{C}}$ and high-frequency feature $\mathbf{F}_{hi}\in\mathbb{R}^{{H}\times{W}\times{C'}}$, we first resize $\mathbf{F}_{hi}$ and obtain $\mathbf{\tilde{F}}_{hi}\in\mathbb{R}^{\hat{H}\times\hat{W}\times{C'}}$. 
Towards highlighting high-frequency characteristics, we employ a gating mechanism to adaptively determine the useful frequency information: 
\begin{equation}
\begin{aligned}
        \hat{\mathbf{F}}^{}_{hi} =\mathbf{\tilde{F}}_{hi}\odot\sigma({\rm DConv}_{3\times 3}(\mathbf{\tilde{F}}_{hi})) ,
\label{eq:F_hi_gate}
\end{aligned}
\end{equation}
where $\hat{\mathbf{F}}^{}_{hi}\in\mathbb{R}^{\hat{H}\times\hat{W}\times{C'}}$ is the processed feature, DConv$_{3\times 3}(\cdot)$ denotes a depth-wise convolution operation with the filter size of 3$\times$3, and $\sigma(\cdot)$ is the GELU activation function~\cite{hendrycks2016gaussian}. 
Then, we leverage the learnable high-frequency prompt components $\mathbf{P}_{hi}\in\mathbb{R}^{\hat{H}\times\hat{W}\times{C'}}$ to make adjustments to the input features, {}{which aims to help distinguish various elements, such as rain patterns and streaks of different orientations and magnitudes in the context of deraining:} 
\begin{equation}
\begin{aligned}
        \mathbf{F}^{prompt}_{hi} =\hat{\mathbf{F}}^{}_{hi}\odot\mathbf{P}_{hi},
\label{eq:F_hi_prompt_channel}
\end{aligned}
\end{equation}
where $\mathbf{F}^{prompt}_{hi}\in\mathbb{R}^{\hat{H}\times\hat{W}\times{C'}}$ is the obtained high-frequency feature prompt. 
%
%

Next, we modify the high-frequency prompt $\mathbf{F}^{prompt}_{hi}$ according to the input feature $\mathbf{F}_l$. 
To be specific, we utilize a depth-wise convolution operator, which acts as a high-pass filter~\cite{park2022how}, to enhance the high-frequency sources in the input $\mathbf{F}^{prompt}_{hi}$. 
Then, we generate $query$ ($\textbf{Q}_{hi}$) projection from $\mathbf{F}_l$, $key$ ($\textbf{K}_{hi}$) and $value$ ($\textbf{V}_{hi}$) projections from the processed feature map $\hat{\mathbf{F}}^{prompt}_{hi} = {\rm DConv_{3
\times 3}}({\mathbf{F}}^{prompt}_{hi})$, respectively. 
%
%
Meanwhile, as the high-frequency information usually corresponds to image details and is a local feature, it could be redundant to calculate global attention. 
%
Therefore, before leveraging the linear layer to obtain the matrices of $\mathbf{Q}_{hi}$, $\mathbf{K}_{hi}$, and $\mathbf{V}_{hi}$, the local window self-attention mechanism is adopted to save computational complexity and capture fine-grained high frequencies, which yields $\textbf{Q}_{hi}=W^{Q_{hi}}_p\cdot R(\mathbf{F}_l)$, $\textbf{K}_{hi}=W^{K_{hi}}_p\cdot R(\hat{\mathbf{F}}^{prompt}_{hi})$, $\textbf{V}_{hi}=W^{V_{hi}}_p\cdot R(\hat{\mathbf{F}}^{prompt}_{hi})$. 
Where $W^{(\cdot)}_p$ represents the projection matrices, and $R(\cdot)$ denotes the window partition strategy~\cite{liu2021swin}. 
Generally, we have $\mathbf{Q}_{hi}\in\mathbb{R}^{\frac{\hat{H}\hat{W}}{M^2}\times{M^2}\times\hat{C}}$, $\mathbf{K}_{hi}\in\mathbb{R}^{\frac{\hat{H}\hat{W}}{M^2}\times\hat{C}\times{M^2}}$, and $\mathbf{V}_{hi}\in\mathbb{R}^{\frac{\hat{H}\hat{W}}{M^2}\times{M^2}\times\hat{C}}$, where $M^2$ is the size of split windows. 
The attention matrix is thus calculated to tune the high-frequency prompt as: 
%
\begin{equation}
\begin{aligned}
        \mathbf{F}^{out}_{hi} = \mathbf{V}_{hi} \cdot {\rm Softmax}(\mathbf{K}_{hi} \cdot \mathbf{Q}_{hi} / \sqrt{d}),
\label{eq:F_hi_prompt_modulate}
\end{aligned}
\end{equation}
where $\mathbf{F}^{out}_{hi}\in\mathbb{R}^{\hat{H}\times\hat{W}\times\hat{C}}$ is the output feature map of the high-frequency prompt modulation branch; ${d}$ is the query/key dimension, following \cite{iccv2021_swinIR}. 

\noindent\textbf{Low-frequency Prompt Modulator.}
Given the two input feature maps, including the $l$-level feature $\mathbf{F}_l\in\mathbb{R}^{\hat{H}\times\hat{W}\times\hat{C}}$ and low-frequency feature $\mathbf{F}_{lo}\in\mathbb{R}^{{H}\times{W}\times{C'}}$, we first resize $\mathbf{F}_{lo}$ and obtain $\mathbf{\tilde{F}}_{lo}\in\mathbb{R}^{\hat{H}\times\hat{W}\times{C'}}$. 
Towards handling low-frequency signals effectively, we project $\mathbf{\tilde{F}}_{lo}$ into the frequency domain via the fast Fourier transform (FFT). 
Then, a gating mechanism is adopted to control the useful low-frequency components flow forward:
\begin{equation}
\begin{aligned}
        \hat{\mathbf{F}}_{lo} =\mathcal{F}(\mathbf{\tilde{F}}_{lo})\odot\sigma({\rm Conv}_{1\times 1}(\mathcal{F}(\mathbf{\tilde{F}}_{lo}))) ,
\label{eq:F_lo_FFT_gate}
\end{aligned}
\end{equation}
where $\hat{\mathbf{F}}_{lo}\in\mathbb{R}^{\hat{H}\times(\frac{\hat{W}}{2}+1)\times{2C'}}$ is the processed feature, $\mathcal{F}(\cdot)$ represents the FFT. 
Next, we calibrate the input features by injecting learnable low-frequency prompt components $\mathbf{P}_{lo}\in\mathbb{R}^{\hat{H}\times(\frac{\hat{W}}{2}+1)\times{2C'}}$, which is then transformed back to the spatial domain:  
\begin{equation}
\begin{aligned}
        \mathbf{F}^{prompt}_{lo} =\mathcal{F}^{-1}(\hat{\mathbf{F}}_{lo}\odot\mathbf{P}_{lo}),
\label{eq:F_lo_prompt_spatial}
\end{aligned}
\end{equation}
where $\mathbf{F}^{prompt}_{lo}\in\mathbb{R}^{\hat{H}\times\hat{W}\times{C'}}$ is the generated low-frequency feature prompt, and $\mathcal{F}^{-1}(\cdot)$ denotes the inverse FFT. 

Noted, we perform the feature transformation in the Fourier domain for efficient global information interaction. 
The convolution theorem~\cite{rabiner1975theory,oppenheim1999discrete} indicates the Hadamard product of two signals in the Fourier domain equals to implement the Fourier transform of a convolution of these two signals in the original spatial domain. 
Base on this insight, we can combine Eq.(\ref{eq:F_lo_FFT_gate}) and Eq.(\ref{eq:F_lo_prompt_spatial}): 
\begin{equation}
\begin{aligned}
        \mathbf{F}^{prompt}_{lo} &=\mathcal{F}^{-1}(\mathcal{F}(\mathbf{\tilde{F}}_{lo})\odot\sigma({\rm Conv}_{1\times 1}(\mathcal{F}(\mathbf{\tilde{F}}_{lo}))) \odot\mathbf{P}_{lo}) \\
        &=\mathcal{F}^{-1}(\mathcal{F}(\mathbf{\tilde{F}}_{lo}\circledast\mathcal{F}^{-1}(\sigma({\rm Conv}_{1\times 1}(\mathcal{F}(\mathbf{\tilde{F}}_{lo}))) \odot\mathbf{P}_{lo}))) \\
        &=\mathbf{\tilde{F}}_{lo}\circledast\mathcal{F}^{-1}(\sigma({\rm Conv}_{1\times 1}(\mathcal{F}(\mathbf{\tilde{F}}_{lo}))) \odot\mathbf{P}_{lo})
\label{eq:inference_lo_convTheorem}
\end{aligned}
\end{equation}
%
where `$\circledast$' is the convolution operation. 
Since $\mathcal{F}^{-1}(\sigma({\rm Conv}_{1\times 1}(\mathcal{F}(\mathbf{\tilde{F}}_{lo})))\odot\mathbf{P}_{lo})$ is a tensor that shares the same shape with $\mathbf{\tilde{F}}_{lo}$, it can be served as a dynamic depth-wise convolution kernel as large as $\mathbf{\tilde{F}}_{lo}$ in spatial domain {}{while introducing less model complexity}. 

Subsequently, we further modulate the low-frequency visual prompt $\mathbf{F}^{prompt}_{lo}$ with the guidance of the input feature $\mathbf{F}_l$. 
Specifically, we adopt an adaptive average pooling operator, which serves as a low-pass filter~\cite{voigtman1986low}, to enhance the low-frequency content in the input $\mathbf{F}^{prompt}_{lo}$. 
After that, we generate $query$ ($\textbf{Q}_{lo}$) projection from reshaped $\mathbf{F}_l$, $key$ ($\textbf{K}_{lo}$) and $value$ ($\textbf{V}_{lo}$) projections from the average-pooled feature $\hat{\mathbf{F}}^{prompt}_{lo} = {\rm AAP}(\mathbf{F}^{prompt}_{lo})$, respectively. 
Here, ${\rm AAP(\cdot)}$ means the adaptive average pooling operation. 
To this end, 1$\times$1 convolution is employed to aggregate pixel-wise cross-channel context, which yields $\textbf{Q}_{lo}=W^{Q_{lo}}_p\mathbf{F}_l$, $\textbf{K}_{lo}=W^{K_{lo}}_p\hat{\mathbf{F}}^{prompt}_{lo}$, $\textbf{V}_{lo}=W^{V_{lo}}_p\hat{\mathbf{F}}^{prompt}_{lo}$. 
Where $W^{(\cdot)}_p$ is the 1$\times$1 point-wise convolution. 
Next, we calculate the dot-product interaction of query and key projections, which generates a transposed-attention map A of size $\mathbb{R}^{\hat{H}\hat{W}\times1}$. 
Overall, the process of modulating the low-frequency prompt is defined as: 
\begin{equation}
\begin{aligned}
        \mathbf{F}^{out}_{lo} = \mathbf{V}_{lo} \cdot {\rm Softmax}(\mathbf{K}_{lo} \cdot \mathbf{Q}_{lo} / \alpha),
\label{eq:F_lo_prompt_modulate}
\end{aligned}
\end{equation}
where $\mathbf{F}^{out}_{lo}\in\mathbb{R}^{\hat{H}\times\hat{W}\times{C'}}$ is the output feature map of the low-frequency prompt modulation branch; $\mathbf{Q}_{lo}\in\mathbb{R}^{\hat{H}\hat{W}\times{C'}}$, $\mathbf{K}_{lo}\in\mathbb{R}^{C'\times{1}}$, and $\mathbf{V}_{lo}\in\mathbb{R}^{1\times{C'}}$ are the input matrices; $\alpha$ is the learnable scaling parameter. 

For both low/high-frequency modulators, we perform the attention map calculation several times in parallel, and these results are then concatenated for multi-head self-attention (MSA)~\cite{vaswani2017attention}. 

%

\section{Experiments}
\label{sec:exp}
In this section, we evaluate the performance of the proposed FPro on removing various degradations, such as rain streak, raindrop, and moir{\'e} pattern. 
Due to limited space, we include more experimental results (\eg, dehazing on SOTS~\cite{li2018benchmarking} and deblurring on GoPro~\cite{cvpr2017_gopro}) and details in the supplemental material.
\subsection{Experimental settings}
\label{ssec:expSet}
\noindent\textbf{Metrics.}
%
We adopt commonly used peak signal-to-noise ratio (PSNR)~\cite{wang2004image} and structural similarity (SSIM) metrics to evaluate restored images. 
Meanwhile, perceptual metric NIQE~\cite{mittal2012making} is employed as a non-reference metric. 
Following previous works~\cite{wang2022uformer,wang2020model}, PSNR/SSIM computations are implemented on the Y channel in the YCbCr space for the image deraining task, while calculated in the RGB color space for other restoration tasks. 
%
%
In the reported tables, the best and second-best scores are \textbf{highlighted} and \underline{underlined}, respectively. 

\begin{table*}[t]\footnotesize
\begin{minipage}{.48\linewidth}
\caption{Quantitative comparison on {SPAD} \cite{wang2019spatial} for rain streak removal.}
\vspace{-2mm}
\centering
\scalebox{0.96}
{
\begin{tabular}{cccccc}
\toprule[0.8pt]
\multicolumn{2}{c}{}  & \multicolumn{4}{c}{\textbf{SPAD}~\cite{wang2019spatial}}\\ 
\multicolumn{2}{l|}{\textbf{Method}}  & \multicolumn{2}{c}{PSNR~$\uparrow$}& \multicolumn{2}{c}{SSIM~$\uparrow$}   \\ \midrule[0.8pt]

\multicolumn{2}{l|}{DDN~\cite{fu2017removing}}        &  \multicolumn{2}{c}{36.16}& \multicolumn{2}{c}{0.9463}  \\
    \multicolumn{2}{l|}{PReNet~\cite{ren2019progressive}}& \multicolumn{2}{c}{40.16}& \multicolumn{2}{c}{0.9816}  \\
\multicolumn{2}{l|}{RCDNet~\cite{wang2020model}}   & \multicolumn{2}{c}{43.36}& \multicolumn{2}{c}{0.9831}  \\

\multicolumn{2}{l|}{MPRNet~\cite{zamir2021multi}}     & \multicolumn{2}{c}{43.64 }& \multicolumn{2}{c}{0.9844}  \\
\multicolumn{2}{l|}{SPAIR~\cite{purohit2021spatially}}                   & \multicolumn{2}{c}{44.10}& \multicolumn{2}{c}{0.9872}  \\

\multicolumn{2}{l|}{Uformer-S~\cite{wang2022uformer}}   & \multicolumn{2}{c}{46.13}& \multicolumn{2}{c}{0.9913}    \\
\multicolumn{2}{l|}{SCD-Former~\cite{Guo_2023_ICCV}}  & \multicolumn{2}{c}{46.89}& \multicolumn{2}{c}{\textbf{0.9941}}    \\
\multicolumn{2}{l|}{IDT~\cite{xiao2022image}}         & \multicolumn{2}{c}{47.34}& \multicolumn{2}{c}{0.9929}    \\

\multicolumn{2}{l|}{Restormer~\cite{zamir2022restormer}}                        & \multicolumn{2}{c}{47.98}& \multicolumn{2}{c}{0.9921}  \\
\multicolumn{2}{l|}{DRSformer~\cite{DRSformer}}       & \multicolumn{2}{c}{\underline{48.53}}& \multicolumn{2}{c}{0.9924}    \\
\midrule[0.8pt]
\multicolumn{2}{l|}{FPro~(Ours)}           & \multicolumn{2}{c}{\textbf{48.99}}& \multicolumn{2}{c}{\underline{0.9936}}  \\



\bottomrule[0.8pt]
\end{tabular}}
\label{tab:derain_all}
\end{minipage}~~~~\begin{minipage}{.48\linewidth}
\caption{Quantitative comparison on AGAN-Data~\cite{qian2018attentive} for raindrop removal.
}
\vspace{-2.mm}
\centering
\scalebox{0.96}
{
\begin{tabular}{ccc}
\toprule[0.8pt]
\multicolumn{1}{l}{}   & \multicolumn{2}{c}{\textbf{AGAN-Data}~\cite{qian2018attentive}
}\\ 
\multicolumn{1}{l|}{\textbf{Method}}&\multicolumn{1}{c}{{PSNR~$\uparrow$}} & \multicolumn{1}{c}{{SSIM~$\uparrow$}}\\ 
\midrule[0.8pt]
\multicolumn{1}{l|}{Eigen's~\cite{eigen2013restoring}} & \multicolumn{1}{c}{21.31}& \multicolumn{1}{c}{0.757} \\
\multicolumn{1}{l|}{Pix2pix~\cite{isola2017image}} & \multicolumn{1}{c}{28.02}& \multicolumn{1}{c}{0.855} \\
\multicolumn{1}{l|}{Uformer-S~\cite{wang2022uformer}}& \multicolumn{1}{c}{29.42}& \multicolumn{1}{c}{0.906} \\
\multicolumn{1}{l|}{WeatherDiff$_{128}$~\cite{ozdenizci2023restoring}}& \multicolumn{1}{c}{29.66}& \multicolumn{1}{c}{0.923} \\
\multicolumn{1}{l|}{TransWeather~\cite{Valanarasu_2022_CVPR}}  & \multicolumn{1}{c}{30.17}& \multicolumn{1}{c}{0.916} \\
\multicolumn{1}{l|}{DuRN~\cite{liu2019dual}}  & \multicolumn{1}{c}{31.24}& \multicolumn{1}{c}{0.926} \\%
\multicolumn{1}{l|}{RaindropAttn~\cite{quan2019deep}} & \multicolumn{1}{c}{31.37}& \multicolumn{1}{c}{0.918} \\
\multicolumn{1}{l|}{AttentiveGAN~\cite{qian2018attentive}} & \multicolumn{1}{c}{31.59}& \multicolumn{1}{c}{0.917} \\
\multicolumn{1}{l|}{IDT~\cite{xiao2022image}} & \multicolumn{1}{c}{{31.63}}& \multicolumn{1}{c}{\underline{0.936}} \\%
\multicolumn{1}{l|}{Restormer~\cite{zamir2022restormer}} & \multicolumn{1}{c}{\underline{31.68}}& \multicolumn{1}{c}{{0.934}} \\
\midrule[0.8pt]
\multicolumn{1}{l|}{{FPro}~(Ours)} & \multicolumn{1}{c}{\textbf{31.96}}& \multicolumn{1}{c}{\textbf{0.937}} \\
\bottomrule
\end{tabular}}
\label{tab:raindrop_compare}
\end{minipage}
\end{table*}
\begin{table}[t]
\caption{Quantitative comparison on TIP-2018~\cite{MSNet} for moir{\'e} pattern removal.
}
\vspace{-3.mm}
\centering
\scalebox{0.750}{
\begin{tabular}{cccccccccccc}
\toprule[0.8pt]
 \multicolumn{1}{c|}{\textbf{Method}}& \multicolumn{1}{c}{AMNet}& \multicolumn{1}{c}{DMCNN} & \multicolumn{1}{c}{UNet} & \multicolumn{1}{c}{WDNet} & \multicolumn{1}{c}{MopNet}& \multicolumn{1}{c}{TAPE-Net}& \multicolumn{1}{c}{FH$\rm D^{2}$eNet} & \multicolumn{1}{c}{MBCNN}&  \multicolumn{1}{c}{Uformer-S}& \multicolumn{1}{c}{Wang \etal} & \multicolumn{1}{|c}{FPro}\\
\multicolumn{1}{c|}{} &{\cite{AMNet_TCSVT21}}& \multicolumn{1}{c}{\cite{MSNet}}& \multicolumn{1}{c}{\cite{ronneberger2015u}} & \multicolumn{1}{c}{\cite{WuNet_eccv20}} & \multicolumn{1}{c}{\cite{MopNet}}& \multicolumn{1}{c}{\cite{TAPE-Net_eccv22}}& \multicolumn{1}{c}{\cite{FHDe2Net_eccv20}} & \multicolumn{1}{c}{\cite{Zheng_2020_CVPR}}& \multicolumn{1}{c}{\cite{wang2022uformer}}& \multicolumn{1}{c}{\cite{wang2023coarse}} & \multicolumn{1}{|c}{(Ours)}\\
\midrule[0.5pt]
\multicolumn{1}{c|}{PSNR~$\uparrow$} &\multicolumn{1}{c}{25.47}& \multicolumn{1}{c}{ 26.10}& \multicolumn{1}{c}{26.49} & \multicolumn{1}{c}{ 27.12} & \multicolumn{1}{c}{27.48}& \multicolumn{1}{c}{27.52}& \multicolumn{1}{c}{27.79} & \multicolumn{1}{c}{28.40} & \multicolumn{1}{c}{28.63}& \multicolumn{1}{c}{\underline{28.87}} & \multicolumn{1}{|c}{\textbf{29.25}}\\
 \multicolumn{1}{c|}{SSIM~$\uparrow$} & \multicolumn{1}{c}{{0.833}}& \multicolumn{1}{c}{0.844}&\multicolumn{1}{c}{0.864} & \multicolumn{1}{c}{ 0.854} & \multicolumn{1}{c}{ 0.861}& \multicolumn{1}{c}{0.866}& \multicolumn{1}{c}{0.867} & \multicolumn{1}{c}{0.871}&  \multicolumn{1}{c}{{0.872}}& \multicolumn{1}{c}{\textbf{0.894}} & \multicolumn{1}{|c}{\underline{0.879}}\\
\bottomrule[0.8pt]
\end{tabular}}
\label{tab:demoire}
\vspace{-1.5mm}
\end{table}
\noindent\textbf{Implementation Details.}
FPro contains $N_1=3$ levels encoder-decoder, where the encode and decoder share the same $N_2$=[2,3,6] blocks. 
We set embedding dimensions $C$ as 48, and the attention heads as [2,4,8]. 
The expanding channel capacity factor in FFN is 3. 
The default split window size in HPM is set as $M=8$.
The pixel-unshuffle and pixel-shuffle are employed for downsampling and upsampling. 
%
We use the AdamW optimizer with the initial learning rate $3e^{-4}$ gradually reduced to $1e^{-6}$ with the cosine annealing to train FPro and adopt the widely used loss function~\cite{wang2023promptrestorer} to constrain the network training.

\begin{figure*}[t]
\tiny
\centering
\begin{tabular}{ccc}
\begin{adjustbox}{valign=t}
\begin{tabular}{cccccc}
\includegraphics[width=0.156\textwidth]{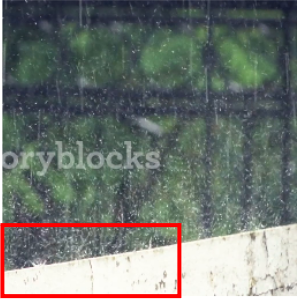}
 &
\includegraphics[width=0.156\textwidth]{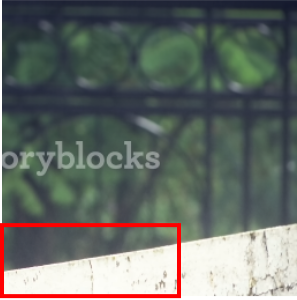}
 &
\includegraphics[width=0.156\textwidth]{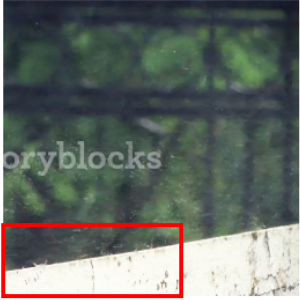}
 &
\includegraphics[width=0.156\textwidth]{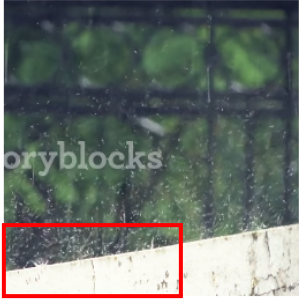}
 &
\includegraphics[width=0.156\textwidth]{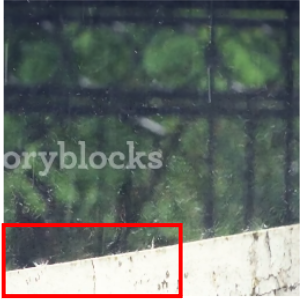}
 &
\includegraphics[width=0.156\textwidth]{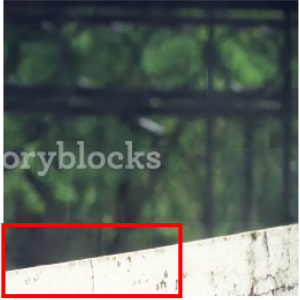}
\\
\includegraphics[width=0.156\textwidth]{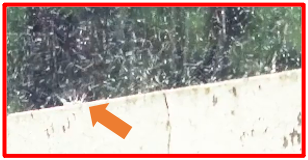}
 &
\includegraphics[width=0.156\textwidth]{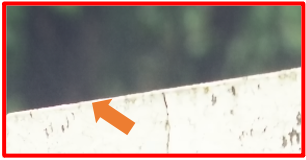}
 &
\includegraphics[width=0.156\textwidth]{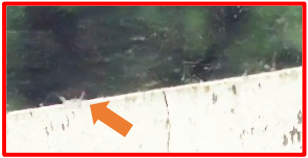}
 &
\includegraphics[width=0.156\textwidth]{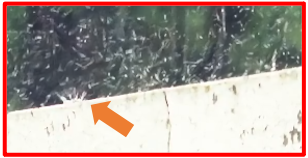}
 &
\includegraphics[width=0.156\textwidth]{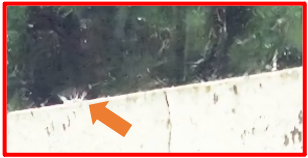}
 &
\includegraphics[width=0.156\textwidth]{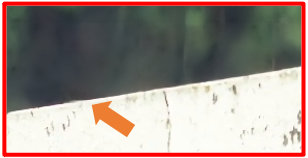}
\\
Rainy  &
Reference  &
DRSformer\cite{DRSformer}  &
RCDNet\cite{wang2020model}  &
Restormer\cite{zamir2022restormer}  &
FPro 
\end{tabular}
\end{adjustbox}
\end{tabular}
\vspace{-2.mm}
\caption{Qualitative comparisons with state-of-the-art methods on SPAD~\cite{wang2019spatial} for real rain removal. (Zoom in for a better view.)}
\label{pic:derain_real}
\vspace{-1.5mm}
\end{figure*}
\begin{figure*}[t]
\tiny
\centering
\begin{tabular}{ccc}
\hspace{-0.1cm}
\begin{adjustbox}{valign=t}
\begin{tabular}{cccccc}
\includegraphics[width=0.156\textwidth]{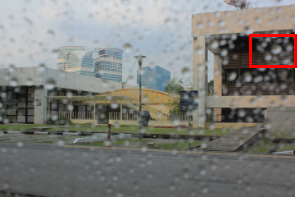}
 &
\includegraphics[width=0.156\textwidth]{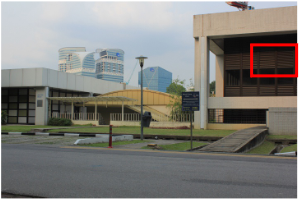}
 &
\includegraphics[width=0.156\textwidth]{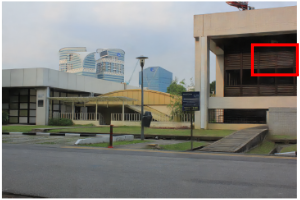}
 &
\includegraphics[width=0.156\textwidth]{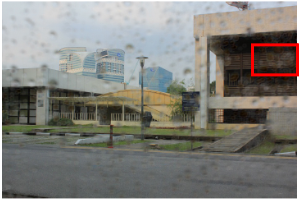}
 &
\includegraphics[width=0.156\textwidth]{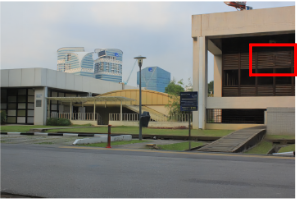}
 &
\includegraphics[width=0.156\textwidth]{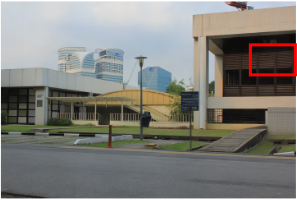}
\\
\includegraphics[width=0.156\textwidth]{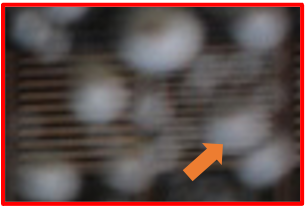}
 &
\includegraphics[width=0.156\textwidth]{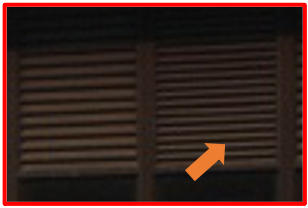}
 &
\includegraphics[width=0.156\textwidth]{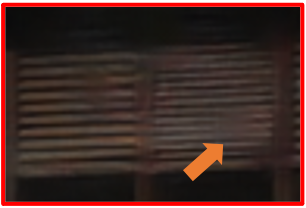}
 &
\includegraphics[width=0.156\textwidth]{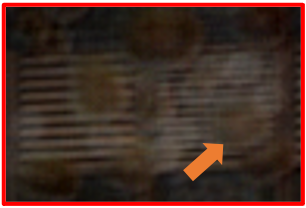}
 &
\includegraphics[width=0.156\textwidth]{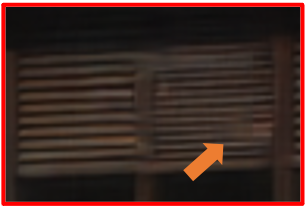}
 &
\includegraphics[width=0.156\textwidth]{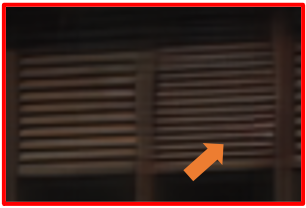}
\\
Raindrop  &
Reference  &
RaindropAttn~\cite{quan2019deep} &
Uformer\cite{wang2022uformer}  &
Restormer\cite{zamir2022restormer}  &
FPro 
\end{tabular}
\end{adjustbox}
\end{tabular}
\vspace{-2.mm}
\caption{Qualitative comparisons with state-of-the-art methods on AGAN-Data~\cite{qian2018attentive} for raindrop removal. (Zoom in for a better view.)}
\label{pic:deraindrop}
\vspace{-5.5mm}
\end{figure*}

\subsection{Main Results}
\noindent\textbf{Rain Streak Removal.} 
We compare the proposed FPro with several general image restoration approches~\cite{purohit2021spatially,zamir2021multi,wang2022uformer,zamir2022restormer} as well as with task-specfic methods~\cite{fu2017removing,ren2019progressive,wang2020model,Guo_2023_ICCV,xiao2022image,DRSformer}. 
Tab.~\ref{tab:derain_all} shows that the proposed FPro makes superior performance against current state-of-the-art methods for real image deraining on SPAD~\cite{wang2019spatial}. 
Compared to the previous best approach DRSformer~\cite{DRSformer}, FPro achieves a 0.46 dB performance boost. 
In addition, FPro obtains 2.1 dB PSNR improvement when compared to the recent model SCD-Former~\cite{Guo_2023_ICCV}. 
Fig.~\ref{pic:derain_real} provides a visual deraining example, where FPro successfully removes the real rain streak while preserving the structural content.

\noindent\textbf{Raindrop Removal.} 
For image deraindrop, we compare FPro with existing state-of-the-art methods, including Eigen's~\cite{eigen2013restoring}, Pix2pix~\cite{isola2017image}, TransWeather~\cite{Valanarasu_2022_CVPR}, Uformer\cite{wang2022uformer}, WeatherDiff$_{128}$ \cite{ozdenizci2023restoring},  DuRN~\cite{liu2019dual}, RaindropAttn~\cite{quan2019deep}, AttentiveGAN~\cite{qian2018attentive}, IDT~\cite{xiao2022image}, and Restormer~\cite{zamir2022restormer}. 
We report the quantitative results on the AGAN-Data~\cite{qian2018attentive} benchmark in Tab.~\ref{tab:raindrop_compare}. 
Our FPro obtains the best performance against all considered methods in terms of both PSNR and SSIM metrics. 
FPro makes a performance gain of 0.28 dB over the previous best method Restormer~\cite{zamir2022restormer}, and 2.3 dB over the recent method WeatherDiff$_{128}$~\cite{ozdenizci2023restoring}. 
Fig.~\ref{pic:deraindrop} shows the visual comparisons, where FPro generates a result with finer details. 

\begin{figure*}[t]
\tiny
\centering
\begin{tabular}{ccc}
\hspace{-0.1cm}
\begin{adjustbox}{valign=t}
\begin{tabular}{cccccc}
\includegraphics[width=0.156\textwidth]{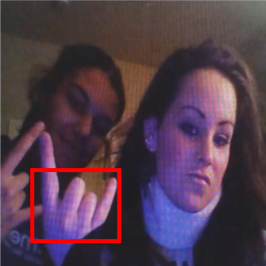}
 &
\includegraphics[width=0.156\textwidth]{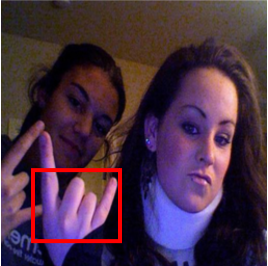}
 &
\includegraphics[width=0.156\textwidth]{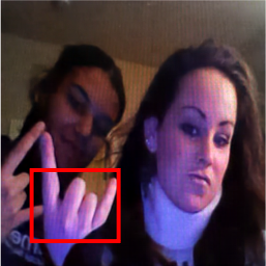}
 &
\includegraphics[width=0.156\textwidth]{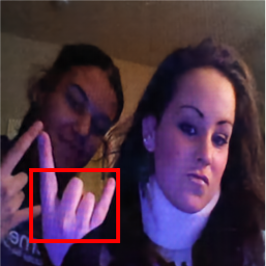}
 &
\includegraphics[width=0.156\textwidth]{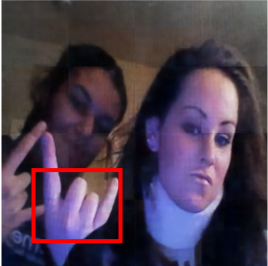}
 &
\includegraphics[width=0.156\textwidth]{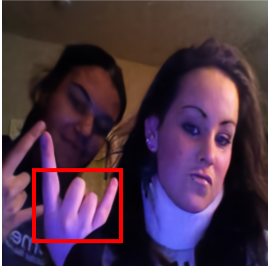}
\\
\includegraphics[width=0.156\textwidth]{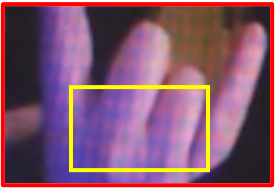}
 &
\includegraphics[width=0.156\textwidth]{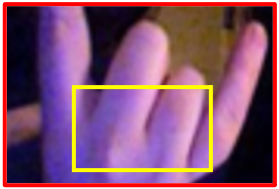}
 &
\includegraphics[width=0.156\textwidth]{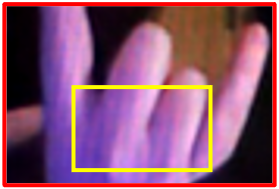}
 &
\includegraphics[width=0.156\textwidth]{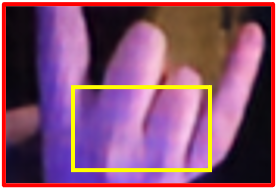}
 &
\includegraphics[width=0.156\textwidth]{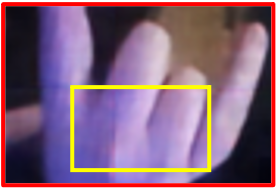}
 &
\includegraphics[width=0.156\textwidth]{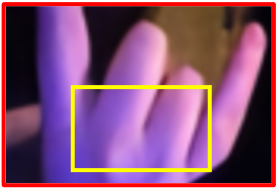}
\\
Moir{\'e}  &
Reference  &
WDNet~\cite{WuNet_eccv20} &
Uformer\cite{wang2022uformer}  &
TAPE\cite{TAPE-Net_eccv22}  &
FPro 
\end{tabular}
\end{adjustbox}
\end{tabular}
\vspace{-3.mm}
\caption{Qualitative comparisons with state-of-the-art methods on TIP-2018~\cite{MSNet} for moir{\'e} pattern removal. (Zoom in for a better view.)
}
\label{pic:demoire}
\vspace{-2.5mm}
\end{figure*}

\noindent\textbf{Moir{\'e} pattern Removal.} 
We conduct moir{\'e} pattern removal experiments on TIP-2018~\cite{MSNet} benchmark, and compare FPro with a wide range of state-of-the-art methods, including AMNet~\cite{AMNet_TCSVT21}, DMCNN~\cite{MSNet}, UNet~\cite{ronneberger2015u}, WDNet~\cite{WuNet_eccv20}, MopNet~\cite{MopNet}, TAPE-Net~\cite{TAPE-Net_eccv22}, FH$\rm D^{2}$eNet~\cite{FHDe2Net_eccv20}, MBCNN~\cite{Zheng_2020_CVPR}, Uformer-S~\cite{wang2022uformer}, and Wang~\etal~\cite{wang2023coarse}. 
In Tab.~\ref{tab:demoire}, FPro yields a 0.38 performance boost against the previous best method Wang~\etal~\cite{wang2023coarse}, and outperforms the recent model TAPE-Net~\cite{TAPE-Net_eccv22} by 1.73 dB in terms of PSNR. 
We present visual comparisons in Fig.~\ref{pic:demoire}, where FPro effectively removes moir{\'e} degradation. 

\begin{table}[b]
\begin{minipage}{.5\linewidth}
\caption{Effectiveness of GDD.}
\vspace{-3mm}
\centering
\scalebox{1}
{
\begin{tabular}{cccc}
\toprule[0.8pt]


\multicolumn{1}{c}{} &\multicolumn{1}{l}{Models}& \multicolumn{1}{|c}{{PSNR}}& \multicolumn{1}{c}{{SSIM}} \\ \midrule[0.8pt]

\multicolumn{1}{c}{(a)} &\multicolumn{1}{l}{Multi DC~\cite{chen2020dynamic}}& \multicolumn{1}{|c}{{48.52}}& \multicolumn{1}{c}{0.9926}   \\
\multicolumn{1}{c}{(b)} &\multicolumn{1}{l}{Multi GDD}& \multicolumn{1}{|c}{{48.91}}& \multicolumn{1}{c}{0.9934} \\
\multicolumn{1}{c}{(c)} &\multicolumn{1}{l}{Single GDD}& \multicolumn{1}{|c}{48.99}& \multicolumn{1}{c}{0.9936}  \\

\bottomrule
\end{tabular}}
\label{tab:abs_edd}
\end{minipage}~~~\begin{minipage}{.45\linewidth}
\caption{Ablation study of DPB. 
}
\vspace{-3.mm}
\centering
\footnotesize
\scalebox{1}
{
\begin{tabular}{cccc}
\toprule[0.8pt]

\multicolumn{1}{c}{} &\multicolumn{1}{l}{Models}& \multicolumn{1}{|c}{{PSNR}}& \multicolumn{1}{c}{{SSIM}} \\ \midrule[0.8pt]

\multicolumn{1}{c}{(a)} &\multicolumn{1}{l}{w/o HPM}& \multicolumn{1}{|c}{{48.77}}& \multicolumn{1}{c}{{0.9931}}   \\
\multicolumn{1}{c}{(b)} &\multicolumn{1}{l}{w/o LPM}& \multicolumn{1}{|c}{{48.89}}& \multicolumn{1}{c}{{0.9933}}   \\
\multicolumn{1}{c}{(c)} &\multicolumn{1}{l}{Full}& \multicolumn{1}{|c}{{48.99}}& \multicolumn{1}{c}{{0.9936}}   \\
\bottomrule
\end{tabular}}
\label{tab:abs_dpb}
\end{minipage}
\end{table}

\begin{figure*}[t]
\begin{minipage}{.32\linewidth}
\tiny
\centering
\begin{tabular}{cc}
\begin{adjustbox}{valign=t}
\begin{tabular}{cc}
\includegraphics[width=0.46\textwidth]{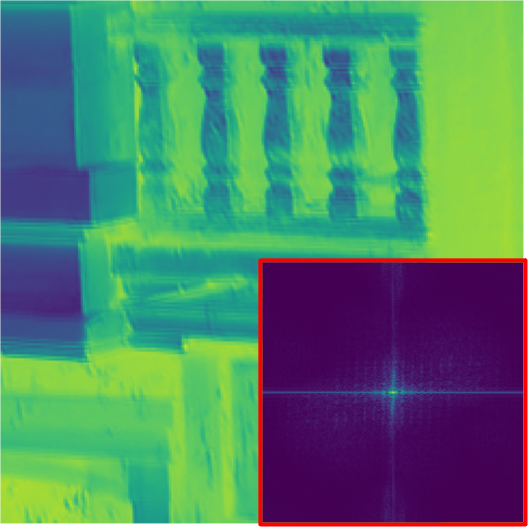} \hspace{-1mm} &
\includegraphics[width=0.46\textwidth]{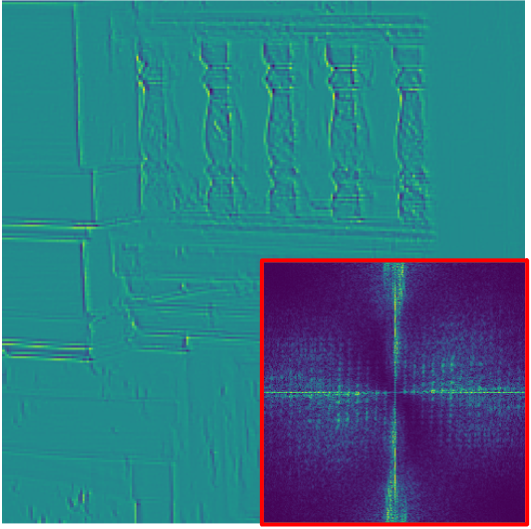} \hspace{-1mm} 
\\
\\
\includegraphics[width=0.45\textwidth]{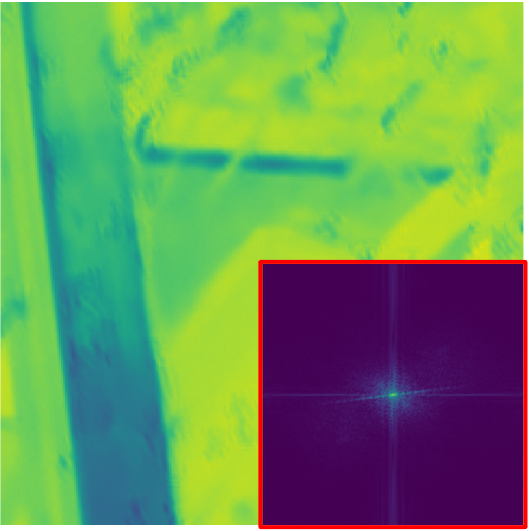} \hspace{-1mm} &
\includegraphics[width=0.45\textwidth]{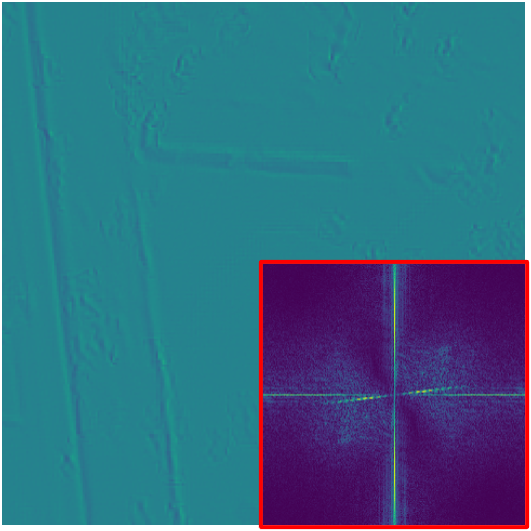} \hspace{-1mm} 
\\ 
(a)Low-frequency \hspace{-1mm} &
(b)High-frequency \hspace{-1mm} 
\\
\end{tabular}
\end{adjustbox}
\end{tabular}
\vspace{-3.mm}
\caption{Feature analysis. 
we visualize the features from the LPM branch (a), and the HPM one (b). 
%
In the right-bottom
, we show the results of the average features over the channel dimension in the Fourier domain. 
(Zoom in for a better view.)
}
\label{pic:visualCom_EDD}
\vspace{-3.5mm}
\end{minipage}~~~~\begin{minipage}{.64\linewidth}
\tiny
\centering
\begin{tabular}{cccc}
\hspace{-0.26cm}
\begin{adjustbox}{valign=t}
\begin{tabular}{cccc}
\includegraphics[width=0.225\textwidth]{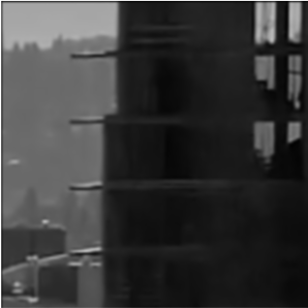} \hspace{-1mm} &
\includegraphics[width=0.225\textwidth]{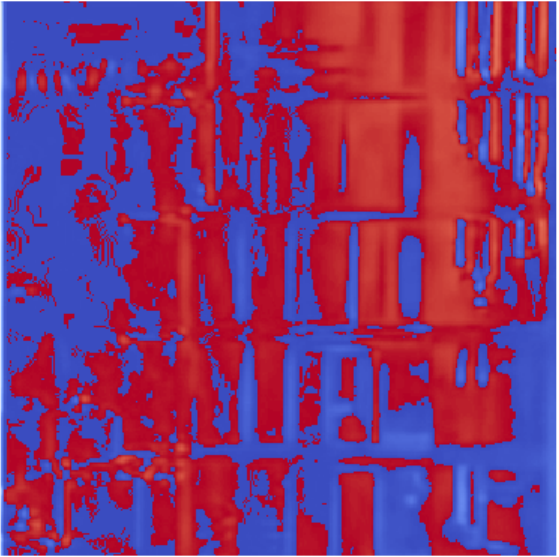} \hspace{-1mm} &
\includegraphics[width=0.225\textwidth]{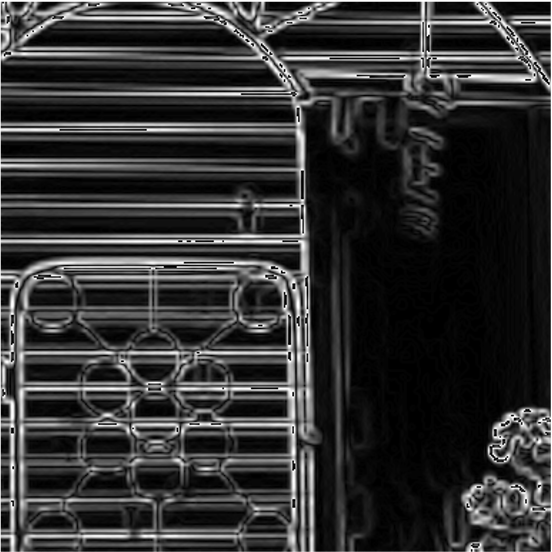} \hspace{-1mm} &
\includegraphics[width=0.225\textwidth]{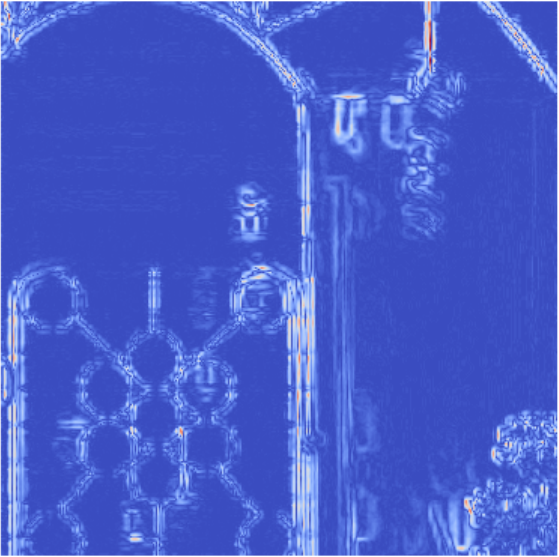} \hspace{-1mm} 
\\
(a) w/o LPM \hspace{-1mm} &
(b) Diff. \hspace{-1mm} &
(c) w/o HPM \hspace{-1mm} &
(d) Diff. \hspace{-1mm} 
\\
\includegraphics[width=0.225\textwidth]{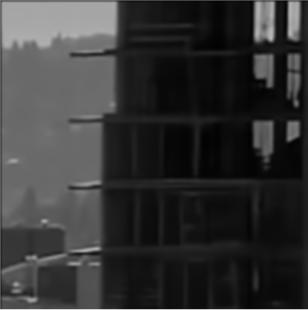} \hspace{-1mm} &
\includegraphics[width=0.225\textwidth]{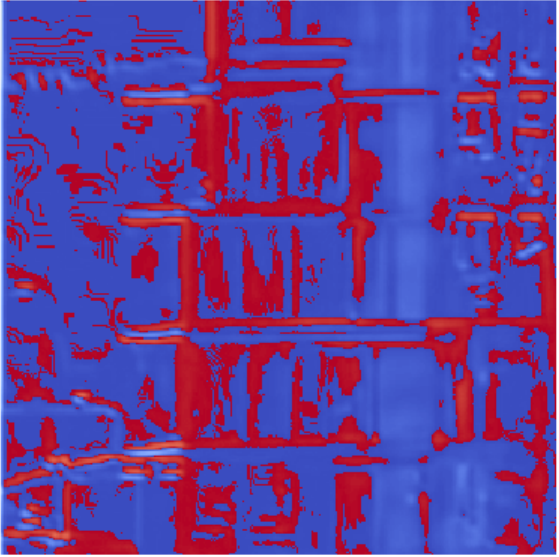} \hspace{-1mm} &
\includegraphics[width=0.225\textwidth]{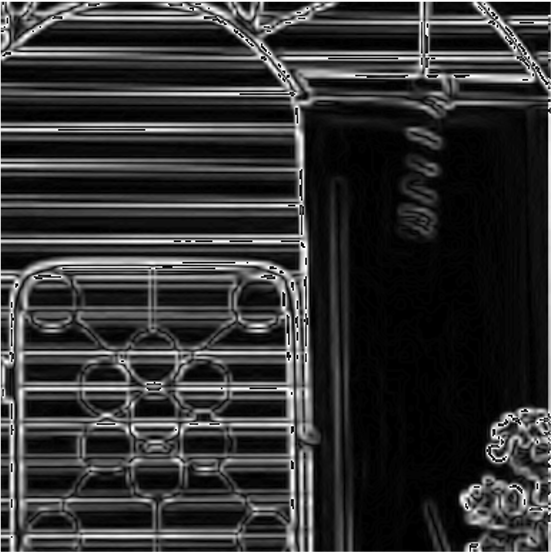} \hspace{-1mm} &
\includegraphics[width=0.225\textwidth]{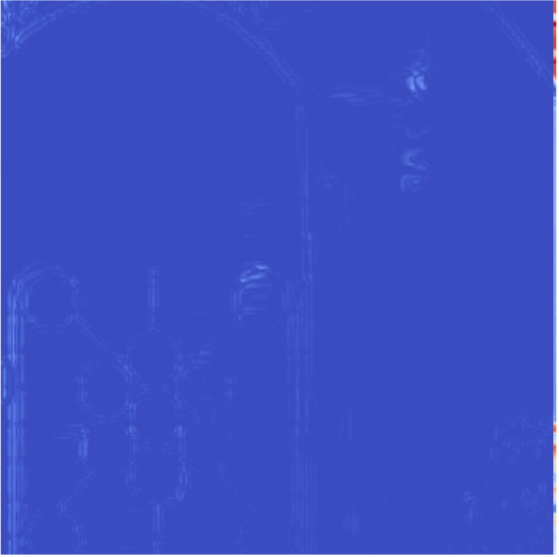} \hspace{-1mm} 
\\ 
(e) w/ LPM \hspace{-1mm} &
(f) Diff. \hspace{-1mm} &
(g) w/ HPM \hspace{-1mm} &
(h) Diff. \hspace{-1mm} 
\\
\end{tabular}
\end{adjustbox}
\end{tabular}
\vspace{-3mm}
\caption{Effect of DBP. 
Columns 1 and 3 show low-pass and high-pass filtered results, while columns 2 and 4 show the difference (Diff.) between processed results with corresponding filtered ground-truth. 
Compared with (a), FPro w/ LPM (e) performs better in capturing information such as structures, resulting in fewer erroneous predictions (f). 
Compared with (c), FPro w/ HPM (g) restores clear edges and shapes, which indicates it enjoys the benefits from the high-frequency information prompt. 
(Zoom in for a better view.)
} 
\label{pic:visualCom_DPB}
\vspace{-3.5mm}
\end{minipage}
\end{figure*}

\subsection{Analysis and Discussion}
For ablation studies, we train deraining models on SPAD~\cite{wang2019spatial} with 256$\times$256 patches for 300K iterations. 
Testing is conducted on SPAD testing dataset~\cite{wang2019spatial}. 
%
%

\noindent\textbf{Effectiveness of Gated Dynamic Decoupler.}
To demonstrate the effectiveness of the Gated Dynamic Decoupler, we conduct experiments on different model variants in Tab.~\ref{tab:abs_edd}. 
Compared to the model equipped with Multiple Dynamic Convolution~\cite{chen2020dynamic} (DC) for separating different frequency parts (a), directly replacing it with GDD (b) results in a performance gain of 0.39 dB in terms of PSNR. 
Meanwhile, instead of injecting GDD into each DPB (b) {}{to employ multiple decouplers}, we attempt to share one GDD module to divide the low-/high frequency information (c), which slightly reduces the complexity {}{(0.02 M)} of the whole framework and brings a 0.08 dB performance boost. 
%

\noindent\textbf{Effectiveness of Dual Propmt Block.}
To investigate the effectiveness of the proposed DPB, we perform an ablation study in Tab.~\ref{tab:abs_dpb} by disabling one core component at a time. 
Our full model achieves the best performance, where disabling HPM or LPM results in a clear drop in performance by 0.22 dB and 0.1 dB, respectively. 
These experimental results demonstrate that both HPM and LPM play a positive role in restoring high-quality images. 
{}{Moreover, we present visualizations to better show the effect of DPB. 
As shown in Fig.~\ref{pic:visualCom_EDD}, we visualize the generated low-/high-frequency feature maps from each branch along with the analysis in the Fourier domain, where the low-frequency prompt feature encodes information such as structures while the high-frequency prompt one focus on information such as edges and texture. 
Meantime, we provide visual comparisons in Fig.~\ref{pic:visualCom_DPB}. 
By prompting the model with low-frequency information, model~(e) performs better in capturing information such as structures and styles, which leads to fewer erroneous predictions (f), compared to baseline model~(a). 
On the other hand, by prompting the model with high-frequency information, model~(g) restores clearer edges and shapes, compared to the baseline model~(c). 
}

\begin{figure*}[t]
\tiny
\centering
\begin{tabular}{ccc}
\begin{adjustbox}{valign=t}
\begin{tabular}{cccccc}
\includegraphics[width=0.16\textwidth]{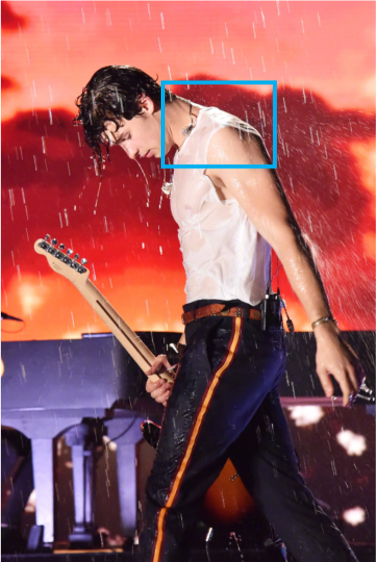}
 &
\includegraphics[width=0.16\textwidth]{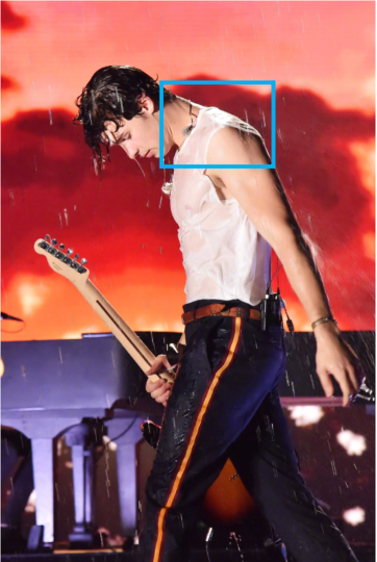}
 &
\includegraphics[width=0.16\textwidth]{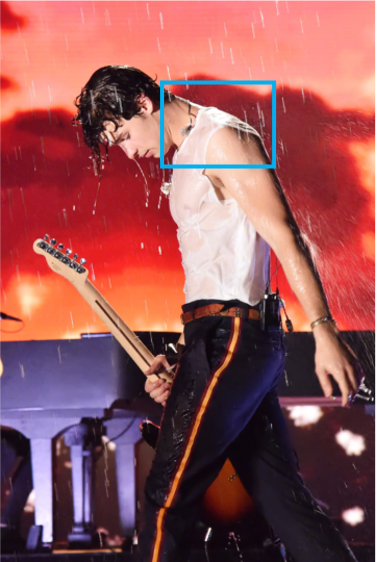}
 &
\includegraphics[width=0.16\textwidth]{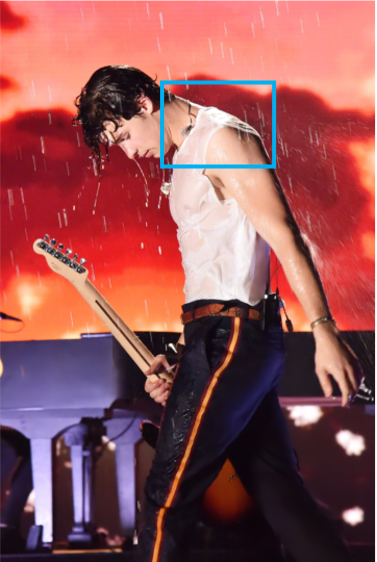}
 &
\includegraphics[width=0.16\textwidth]{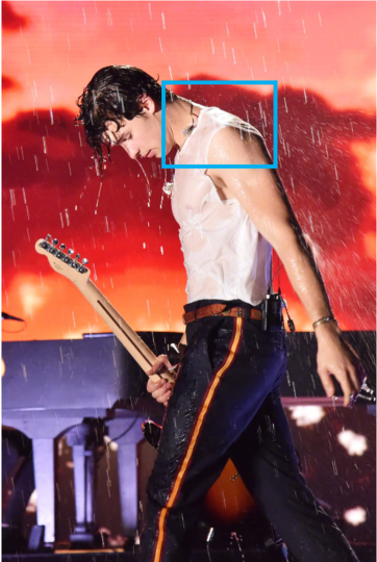}
 &
\includegraphics[width=0.16\textwidth]{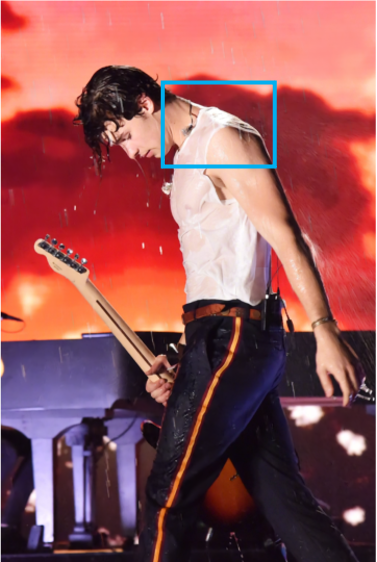}
\\
\includegraphics[width=0.16\textwidth]{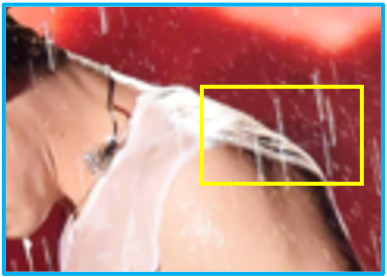}
 &
\includegraphics[width=0.16\textwidth]{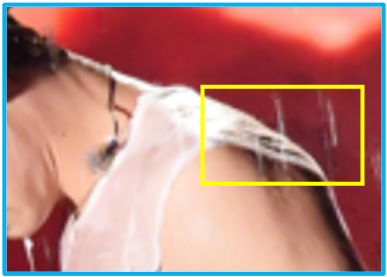}
 &
\includegraphics[width=0.16\textwidth]{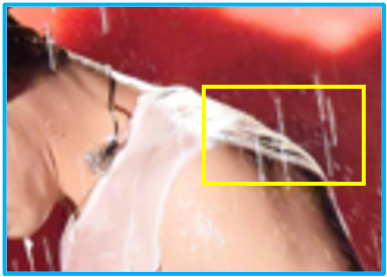}
 &
\includegraphics[width=0.16\textwidth]{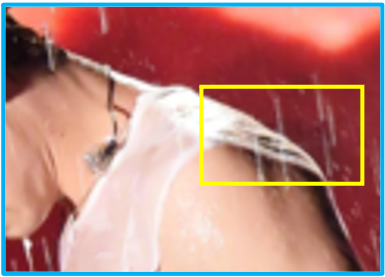}
 &
\includegraphics[width=0.16\textwidth]{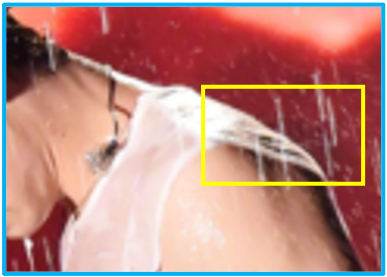}
 &
\includegraphics[width=0.16\textwidth]{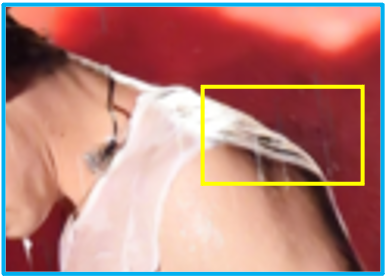}
\\
Rainy  &
Restormer\cite{zamir2022restormer}  &
Uformer\cite{wang2022uformer}  &
IDT\cite{xiao2022image}  &
DRSformer\cite{DRSformer}  &
FPro 
\end{tabular}
\end{adjustbox}
\end{tabular}
\vspace{-3.mm}
\caption{Qualitative comparisons with state-of-the-art methods on Internet-Data~\cite{wang2019spatial} for real rain removal. (Zoom in for a better view.)}
\label{pic:derain_real_niqe}
\vspace{-2.5mm}
\end{figure*}

\noindent\textbf{Perceptual Quality Assessment.}
To test the perceptual quality of the proposed FPro, following~\cite{DRSformer}, we randomly choose 20 rainy images under real-world scenes from Internet-Data~\cite{wang2019spatial} to perform the evaluation. 
As shown in Tab.~\ref{tab:abs_NIQE}, compared to other considered methods, FPro achieves a lower NIQE score, which means the generated results contain clearer content and better perceptual quality. 
Through qualitative comparison in Fig.~\ref{pic:derain_real_niqe}, FPro obtains a visually pleasant result against other models, indicating that it handles unseen degradation well. 

\begin{table}[t]\footnotesize
\caption{Results of no-reference metric NIQE on real-world rainy images. }
\vspace{-3mm}
\centering
\footnotesize
\scalebox{1.06}
{
\begin{tabular}{cccccccccccccccc}
\toprule[0.8pt]
\multicolumn{2}{l|}{{Methods}} & \multicolumn{2}{c}{Input}   & \multicolumn{2}{c}{Uformer-S~\cite{wang2022uformer} } & \multicolumn{2}{c}{Restormer}~\cite{zamir2022restormer} & \multicolumn{2}{c}{IDT~\cite{xiao2022image}} &\multicolumn{2}{c}{DRSformer~\cite{DRSformer}}&\multicolumn{2}{|c}{FPro} \\ 
\midrule[0.8pt]
\multicolumn{2}{l|}{NIQE $\downarrow$} &\multicolumn{2}{c}{5.8012}   & \multicolumn{2}{c}{5.6971}  & \multicolumn{2}{c}{5.6631} & \multicolumn{2}{c}{5.6085} & \multicolumn{2}{c}{5.5942}& \multicolumn{2}{|c}{\textbf{5.2999}}    \\
\bottomrule[0.8pt]
\end{tabular}}
\label{tab:abs_NIQE}
\vspace{-5.5mm}
\end{table}

\begin{table*}[t]
\caption{Model efficiency analysis on SPAD~\cite{wang2019spatial}.}
\vspace{-3.mm}
\centering
\scalebox{.8301}
{
\begin{tabular}{cccccccc}
\toprule[0.8pt]
\multicolumn{1}{c|}{{Method}} & \multicolumn{1}{c}{MPRNet~\cite{zamir2021multi}}& \multicolumn{1}{c}{SwinIR~\cite{iccv2021_swinIR}} & \multicolumn{1}{c}{Uformer-S~\cite{wang2022uformer}} & \multicolumn{1}{c}{Restormer~\cite{zamir2022restormer}}& \multicolumn{1}{c}{IDT~\cite{xiao2022image}} & \multicolumn{1}{c|}{DRSformer~\cite{DRSformer}}& \multicolumn{1}{c}{FPro}
\\ 
\midrule[0.8pt]
\multicolumn{1}{c|}{FLOPs/G} & \multicolumn{1}{c}{175.8} & \multicolumn{1}{c}{238.0} & \multicolumn{1}{c}{\textbf{43.9}} & \multicolumn{1}{c}{174.7}& \multicolumn{1}{c}{\underline{61.9}}& \multicolumn{1}{c|}{242.9}& \multicolumn{1}{c}{{81.9}}%
\\
\multicolumn{1}{c|}{Parameters/M} & \multicolumn{1}{c}{{20.1}} & \multicolumn{1}{c}{\textbf{11.5}} & \multicolumn{1}{c}{20.6} & \multicolumn{1}{c}{26.1}& \multicolumn{1}{c}{\underline{16.4}}& \multicolumn{1}{c|}{33.7}& \multicolumn{1}{c}{22.3}
\\
\multicolumn{1}{c|}{Run-times/s}& \multicolumn{1}{c}{\textbf{0.03}} & \multicolumn{1}{c}{1.83}  & \multicolumn{1}{c}{0.12}&\multicolumn{1}{c}{0.14}&\multicolumn{1}{c}{0.28}&\multicolumn{1}{c|}{\underline{0.08}}&\multicolumn{1}{c}{\underline{0.08}}
\\
\multicolumn{1}{c|}{PSNR/dB}& \multicolumn{1}{c}{43.64} & \multicolumn{1}{c}{44.97}  & \multicolumn{1}{c}{46.13}&\multicolumn{1}{c}{47.98}&\multicolumn{1}{c}{47.34}&\multicolumn{1}{c|}{\underline{48.53}}&\multicolumn{1}{c}{\textbf{48.99}}%
\\
\bottomrule[0.8pt]
\end{tabular}}
\label{tab:efficiency_compare}
\end{table*}
\begin{figure}[t]
\scriptsize
\centering
\begin{tabular}{ccc}
\begin{adjustbox}{valign=t}
\begin{tabular}{cc}
\includegraphics[width=0.5\textwidth]{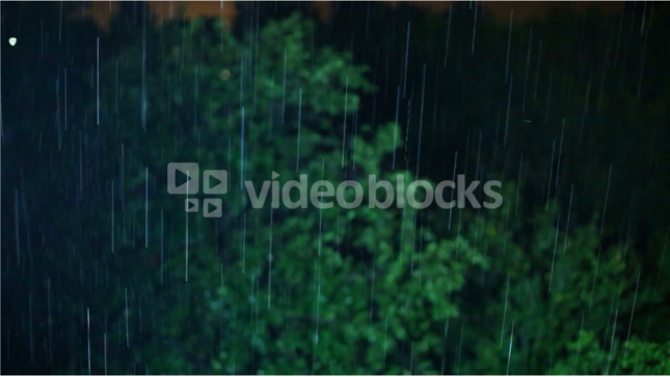}  \hspace{-.2mm} &
\includegraphics[width=0.5\textwidth]{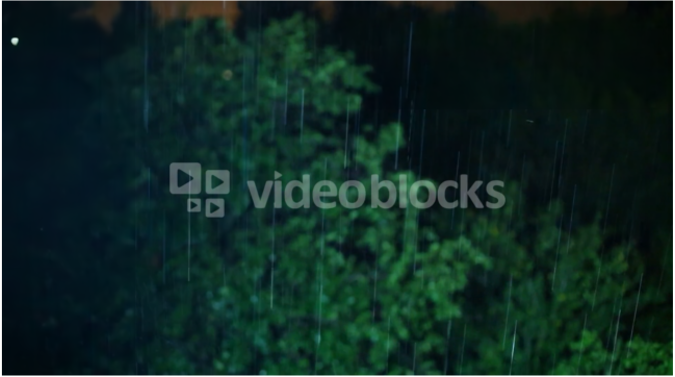} \hspace{-.2mm}  
\\
(a) Input \hspace{-.2mm} &
(b) FPro \hspace{-.2mm} 
\\
\end{tabular}
\end{adjustbox}
\end{tabular}
\vspace{-3.5mm}
\caption{Examples of erroneous restorations. 
Typical failure of FPro can be attributed to heavy degradation in the nighttime real-world scene. 
(Zoom in for a better view.)
} 
\label{pic:fail_case}
\vspace{-3.5mm}
\end{figure}

\noindent\textbf{Model Efficiency.}
We provide the comparison of performance~(PSNR), complexity~(FLOPs and Parameters), and latency~(Run-times) for image deraining. 
FLOPs and Runtimes are measured when input with the size of 256$\times$256, and PSNR scores are tested on SPAD~\cite{wang2019spatial}. 
As shown in Tab.~\ref{tab:efficiency_compare}, though FPro achieves better performance in terms of PSNR metric, it has less model complexity than Restormer~\cite{zamir2022restormer} and DRSformer~\cite{DRSformer}. 
Compared to other CNN-/Transformer-based methods, FPro still has a less or comparable model complexity. 

\begin{wraptable}{r}{0.48\linewidth}
\centering
\vspace{-12.mm}
\caption{Comparisons with alternatives to FPro on Rain100L~\cite{yang2019joint} for deraining.}
\scalebox{.86}{
\begin{tabular}{cccc}
\toprule[0.8pt]
\multicolumn{1}{l}{Models}& \multicolumn{1}{|c}{{Params}}& \multicolumn{1}{c}{{FLOPs}}&  \multicolumn{1}{c}{{PSNR}}  \\ \midrule[0.8pt]

\multicolumn{1}{l}{PromptIR~\cite{potlapalli2023promptir}}& \multicolumn{1}{|c}{{35.6}}& \multicolumn{1}{c}{{173}}&  \multicolumn{1}{c}{37.04}   \\
\multicolumn{1}{l}{PromptRestorer~\cite{wang2023promptrestorer}}& \multicolumn{1}{|c}{24.4}& \multicolumn{1}{c}{186}&  \multicolumn{1}{c}{39.04}   \\
\midrule[0.8pt]
\multicolumn{1}{l}{FPro}& \multicolumn{1}{|c}{{22.3}}& \multicolumn{1}{c}{{82}}&  \multicolumn{1}{c}{39.20}   \\
\bottomrule
\end{tabular}}
\label{tab:abs_compare_PromptIR}
\vspace{-8mm}
\end{wraptable}
\subsection{Comparisons with Alternatives to FPro}
To further demonstrate the superiority of FPro, we compare it with recent prompt-based methods that mine spatial relations as prompts, including PromptIR~\cite{potlapalli2023promptir} and PromptRestorer~\cite{wang2023promptrestorer}. 
As shown in Tab.~\ref{tab:abs_compare_PromptIR}, following 
PromptIR~\cite{potlapalli2023promptir}, we train and validate FPro on Rain100L~\cite{yang2019joint}. 
We achieve a substantial performance gain of 2.16 dB over PromptIR, and a 0.16 dB performance boost against PromptRestorer\footnote{As the code of PromptRestorer is not available for now, we refer to the results of their paper, where the model using additional training data. }. 

\section{Conclusion}
In this paper, we investigated the benefits of prompt learning from a frequency perspective for the task of image restoration. 
%
%
We study two design choices for the exploration of useful frequency characteristics. 
First, when dynamic decoupling the input features with a gating mechanism to select representative elements, we obtain the related frequency components with regard to the specific degradation removal task. 
Then, we propose modulating the low-/high-frequency signals with separate branches, which concern the intrinsic characteristics of feature maps from different frequency bands. 
With these modules, our proposed FPro surpasses previous state-of-the-art methods in several image restoration tasks, while performing competitively in terms of computational cost. 

\noindent\textbf{Limitations.} There remain many avenues for future work and further improvements. 
%
{}{For instance, one could achieve better performance by addressing failure cases are shown in Fig.~\ref{pic:fail_case}, where FPro meets challenges in dealing with heavy degradation in the nighttime real-world scene. 
Intuitively, collecting a large-scale real-world dataset is a potential direction for improvements.}
\bibliographystyle{splncs04}
\bibliography{FPro}
\end{document}